\newcommand{\ARXIV}{} 

\ifdefined\ARXIV
  \documentclass[letterpaper, 10pt, twocolumn]{article}
  \usepackage[margin=0.75in]{geometry}
  \usepackage{libertine}
  \renewcommand\footnotemark{} 
  \date{}
\else
  \documentclass[letterpaper, 10 pt, conference]{ieeeconf}
  \IEEEoverridecommandlockouts
\fi

\usepackage{graphicx}
\usepackage{subfig}
\usepackage{amsmath,amssymb,amsfonts}
\usepackage{algorithm2e}
\usepackage{xcolor}
\usepackage[colorlinks=true, linkcolor=black, urlcolor=cyan, filecolor=black, citecolor=black]{hyperref}
\usepackage{url}
\usepackage{amsmath,bm}
\usepackage{comment}

\usepackage{cite}

\begin{document}

\title{\LARGE \bf Model Predictive Control with Gaussian Processes for Flexible Multi-Modal Physical Human Robot Interaction
\thanks{K. Haninger and C. Hegeler are with the Department of Automation at Fraunhofer IPK, Berlin, Germany. L. Peternel is with the Cognitive Robotics Department, Delft Univerity of Technology, Delft, The Netherlands. Corresponding email: \texttt{kevin.haninger@ipk.fraunhofer.de}}

\thanks{This project has received funding from the European Union's Horizon 2020 research and innovation programme under grant agreement No  820689 — SHERLOCK.}}

\author{Kevin Haninger, Christian Hegeler, and Luka Peternel}
\maketitle

\begin{abstract}
Physical human-robot interaction can improve human ergonomics, task efficiency, and the flexibility of automation, but often requires application-specific methods to detect human state and determine robot response. At the same time, many potential human-robot interaction tasks involve discrete modes, such as phases of a task or multiple possible goals, where each mode has a distinct objective and human behavior. In this paper, we propose a novel method for multi-modal physical human-robot interaction that builds a Gaussian process model for human force in each mode of a collaborative task. These models are then used for Bayesian inference of the mode, and to determine robot reactions through model predictive control. This approach enables optimization of robot trajectory based on the belief of human intent, while considering robot impedance and human joint configuration, according to ergonomic- and/or task-related objectives. The proposed method reduces programming time and complexity, requiring only a low number of demonstrations (here, three per mode) and a mode-specific objective function to commission a flexible online human-robot collaboration task. We validate the method with experiments on an admittance-controlled industrial robot, performing a collaborative assembly task with two modes where assistance is provided in full six degrees of freedom. It is shown that the developed algorithm robustly re-plans to changes in intent or robot initial position, achieving online control at 15 Hz.
\end{abstract}

\section{Introduction}
Facilitated by new compliant manipulators, robots are gradually moving from caged cells separated from humans into production environments involving mixed human-robot teams. One of the crucial elements to enable such collaboration is the ability to control a safe and meaningful physical human-robot interaction (HRI) in a dynamic working environment. Such collaboration can improve ergonomics and efficiency but, today, most industrial applications of physical HRI are offline position teaching and workspace sharing with parallel tasks. While safety challenges contribute to this gap, the complexity of implementing flexible control of physical HRI still remain one of the key issues.
\begin{figure}
    \centering
    \includegraphics[width = 0.8\columnwidth]{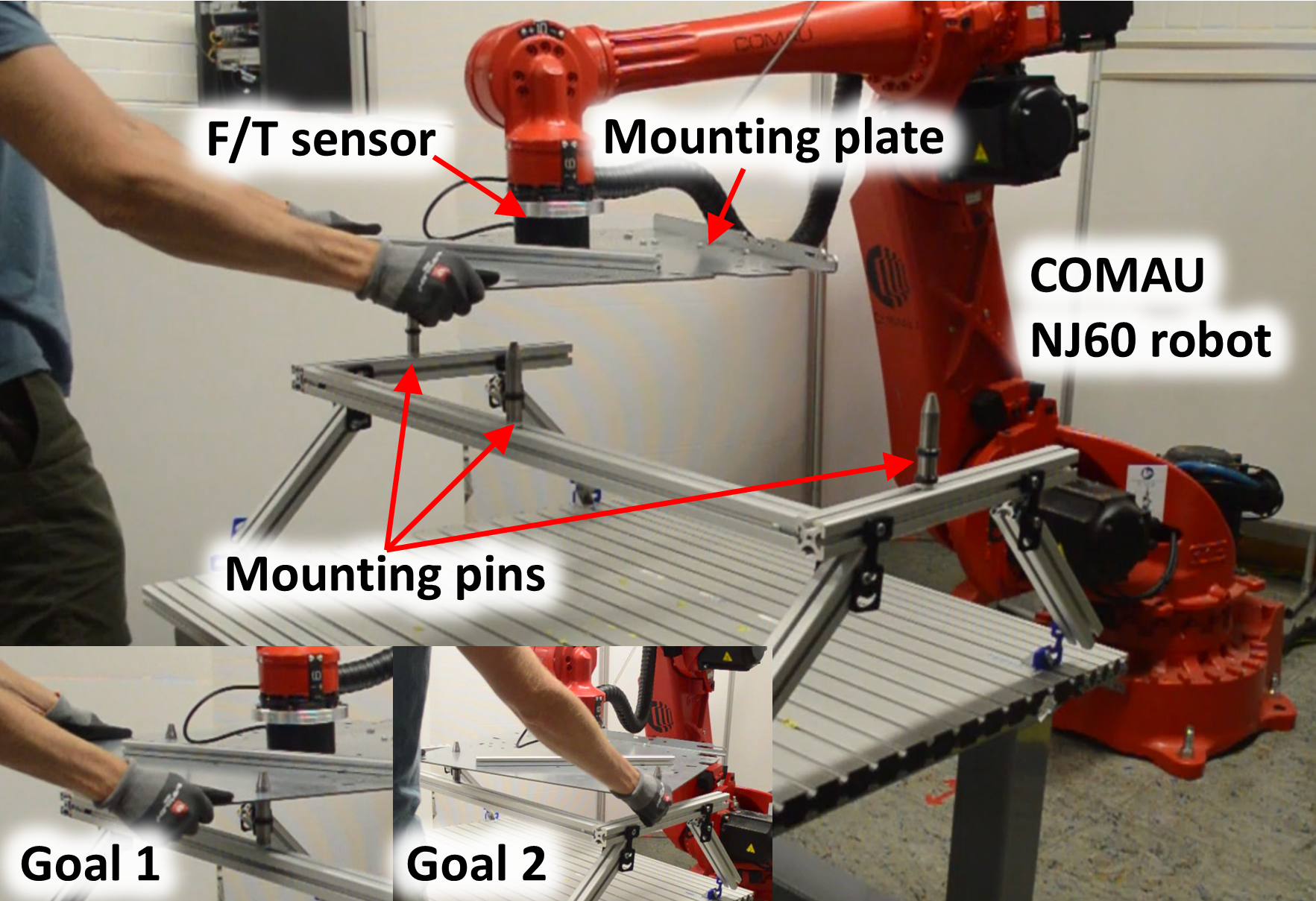}
    \caption{Experimental setup for human-robot collaborative assembly. The task involved the plate to be mounted on one of two sets of pins on the base frame. Pictures in left bottom corner show two possible modes represented by two different assembly goal locations (i.e., Goal 1 and Goal 2). This emulates tasks where the same part is mounted in different locations, such as tires on a car.}
    \label{fig:exp_setup}
    \vspace{-4mm}
\end{figure}

Several recent works in physical HRI have made an important step to account for this gap. For example, adaption of a collaborative robot to the human state can significantly improve task-related performance \cite{takagi2017}, ergonomics \cite{peternel2018}, and user satisfaction \cite{corteville2007}. To do so, the robot should be able to perceive the human state, which presents challenges for both design of sensing \cite{ajoudani2017} and inference methods \cite{jain2018}. On the other hand, some human states or preferences can be more generally modelled, such as the desired motion of the robot/payload \cite{demiris2007, li2014, wang2017, kang2019}, physical fatigue \cite{peternel2018}, and preferred interaction force range \cite{gopinathan2017}. 

Nevertheless, a major part of prompt robot adaption to the human is its ability to infer the human intent during the collaborative task execution \cite{benamor2014,peternel2018}. While human intent is often considered as a continuous variable \cite{takagi2017, kang2019}, it can also reflect discrete changes in the task \cite{khoramshahi2019}, for example a collection of possible goals. Discrete human state has been considered for virtual fixtures \cite{raiola2018}, Dynamic Movement Primitives (DMPs) \cite{khoramshahi2019}, and impedance control \cite{rozo2016}. 

When the robot has the knowledge of human state and intent, it must respond promptly with appropriate actions to facilitate the collaborative task execution. This typically means generation of motion trajectories and often simultaneous impedance adjustments in order to govern a proper interaction behaviour. The trajectories can be chosen to optimize task-related objectives: reducing trajectory jerk \cite{maeda2001, dimeas2015}, minimize positioning error \cite{yang2011a, gribovskaya2011, lee2012}, or render appropriate velocity response \cite{tsumugiwa2002,duchaine2012}. In addition, human operator-related objectives can optimised as well, such as minimizing interaction forces \cite{lamy2009}, or metabolic cost \cite{koller2016}. A common rule is that the robot can take over aspects of of the task that require precision, while the human can handle adaption variations. The `minimum intervention principle' \cite{medina2012} follows this rule, where uncertain degrees of freedom (DOF) should have a lower stiffness \cite{calinon2014, pignat2017}, which can also be interpreted as a risk-sensitive control. Nevertheless, such an approach is not reasonable for all interactive tasks that involve physical constraints \cite{peternel2018a}.

The problem of designing robot response -- including over a belief of human state -- can be simplified by using Model Predictive Control (MPC). MPC can optimize a range of variables, such as motion trajectory \cite{poignet2000nonlinear,norouzzadeh2012towards,rahman2017mpc}, impedance parameters \cite{roveda} and human joint pose, in order to improve a range of objectives that consider ergonomics, task performance, and uncertainty. Advances in optimization solvers, processing power, and toolboxes has increased the use of MPC in many control-engineering domains \cite{lee2011model}, including robotic locomotion \cite{kuindersma2016}, manipulation \cite{rubagotti2019} and rehabilitation \cite{teramae2017emg}. MPC can also accommodate constraints in state or control, allowing some safety or acceptance characteristics to be directly transcribed. However, MPC requires a model, and human models are challenging; a priori human models are limited and collecting human data is expensive. Furthermore, MPC on nonlinear systems has no intrinsic timing or stability guarantees \cite[\S 7.3]{langaker2018}.

MPC has been applied in non-physical HRI for collision avoidance, treating human motion as a disturbance \cite{poignet2000nonlinear}, or with a basic human model \cite{oleinikov2021safety}. While MPC has also been applied in physical HRI on robots with one or two DOF \cite{norouzzadeh2012towards,rahman2017mpc}, these methods only planned a motion with safety constraints according to robot dynamics and did not model the human. MPC with human models in physical HRI has been explored for one DOF tasks in \cite{roveda}. There, the impedance stiffness/damping were adapted with a neural-network based human model and the sampling-based Cross-Entropy Method optimization was employed for MPC. 

In this work, we consider physical HRI tasks which are a discrete collection of modes, each with a distinct objective and human dynamics. We model the human force as a function of mode and robot position using Gaussian Processes (GP) regression and then employ them for:
\begin{enumerate}
    \item a mode inference system, which uses the human models to estimate the current mode, possibly including prior probabilities or transitions for the modes.
    \item a Model Predictive Controller to determine robot trajectory (and other control parameters) according to task- and human-related objectives, evaluated over the belief in mode.
\end{enumerate}

This paper goes beyond the state of the art in physical HRI by allowing a general objective function with a belief over discrete modes, instead of weighting the action associated with each mode by the belief that the mode is currently true, as proposed in \cite{rozo2016, raiola2018, khoramshahi2019}. Compared to the state of the art MPC-based methods for physical HRI \cite{roveda}, this approach considers discrete modes, uses a deterministic optimization-based (not a stochastic sampling-based) MPC, provides 6-DOF assistance, and requires an order of magnitude fewer demonstrations. 

While the proposed approach is general and flexible, a major challenge is computational efficiency. We devote special attention to computational techniques and approximations that improve the computational efficiency. The paper is structured in a following manner: we first introduce the models for the robot and human, the inference and MPC problem formulation, then we describe the proposed approach and implementation.  Experimental validation is done on a medium-payload industrial robot, performing a collaborative assembly task with a human (see Fig. \ref{fig:exp_setup}).

\section{Human and Robot Models}
\label{sec:models}
This section introduces the robot control architecture, robot dynamic model, and human modelling approach. The robot and human dynamics, when coupled, form the stochastic dynamics that is considered in the MPC problem. 

\subsection{Robot Admittance Control}
\begin{figure}
    \centering
    \includegraphics[width=\columnwidth]{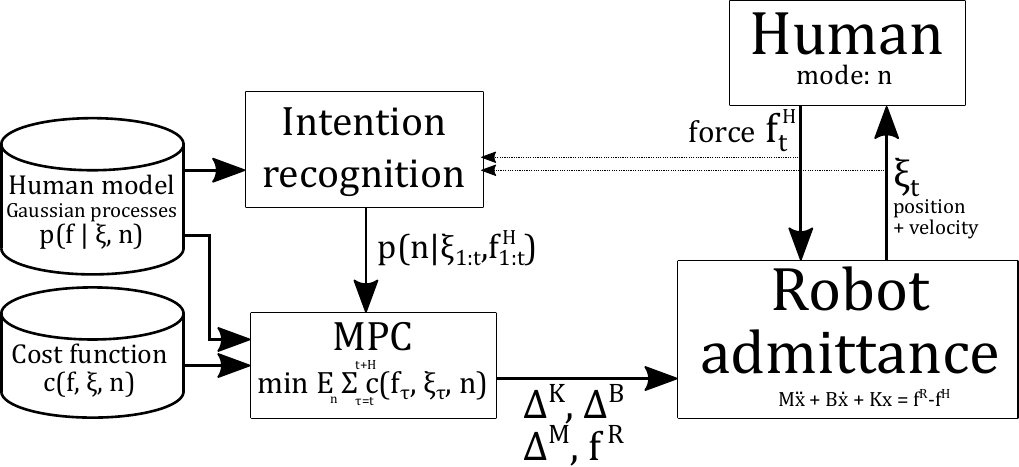}
    \caption{Framework overview. The intention of the human and mode of operation is inferred by using human model encoded by Guassian Processes and real-time measurements. The appropriate collaborative behaviour of the robot is optimised and governed by model predictive control based on the identified modes and corresponding cost functions.}
    \label{fig:architecture}
    \vspace{-4mm}
\end{figure}

As nonlinear MPC does not have intrinsic timing or stability guarantees, the proposed control architecture moves safety concerns, where possible, to a lower-level real-time controller. An admittance controller is used to allow the human or environmental forces to move the robot between MPC updates, which reduces the risk of excessive contact forces. The MPC controller then sets a virtual desired force, and (optionally) changes in the admittance parameters, as seen in Fig. \ref{fig:architecture}. 

A Cartesian admittance controller is realized in the robot's tool center point (TCP) frame to approximate the continuous-time dynamics of
\begin{equation} \label{eq:SysDynCont}
\bm{f}^R - \bm{f}^H = \bm{M}\Ddot{\bm{x}} + \bm{D}\Dot{\bm{x}} + \bm{K}(\bm{x}-\bm{x}_0),
\end{equation}
with pose $\bm{x}\in\mathbb{R}^6$, velocity $\Dot{\bm{x}}\in\mathbb{R}^6$, acceleration $\Ddot{\bm{x}}\in\mathbb{R}^6$, human force $\bm{f}^H\in\mathbb{R}^6$ and virtual desired forces $\bm{f}^R$. The impedance parameters are encoded by the inertia $\bm{M}\in\mathbb{R}^{6\times 6}$, damping $\bm{D}\in\mathbb{R}^{6\times 6}$, and stiffness $\bm{K}\in\mathbb{R}^{6\times 6}$ matrices, which are all diagonal. The rotational elements of $\bm{x}$ and $\bm{f}$ are the angles and torques, respectively, about the three axes of the TCP frame. 

To formulate an MPC problem, the dynamics in \eqref{eq:SysDynCont} have to be discretized. Denote sample time $T_s$ and let subscript $t$ denote the value at time $t_0 + t T_s$, where $t_0$ is the start time at $t=0$. Taking a first-order Euler discretization where $\bm{\xi} = [\bm{x}^T, \dot{\bm{x}}^T]^T$, we derive
\begin{equation}
\begin{aligned}\label{eq:FirstOrderDiscSysDyn}
\bm{\xi}_{t+1} &= \begin{bmatrix} \bm{I} && T_s \bm{I} \\ -T_s \bm{M}^{-1}\bm{K} && \bm{I} - T_s\bm{M}^{-1}\bm{D}\end{bmatrix}
\bm{\xi}_t \\
& \qquad + \begin{bmatrix} \bm{0} \\ T_s\bm{M}^{-1} \end{bmatrix}(\bm{f}_t^H-\bm{f}_t^R),
\end{aligned}
\end{equation}
where $\bm{I}\in\mathbb{R}^{6\times 6}$ and $\bm{0}\in\mathbb{R}^{6\times 6}$ are the identity and zero matrix, respectively.

Using an explicit Euler integrator defined in \eqref{eq:FirstOrderDiscSysDyn} results in oscillation of the discrete state trajectory when $1-T_s\frac{D_{i,i}}{M_{i,i}}<0$, where $\cdot_{i,j}$ denotes the $i,j$-th element of a matrix. This oscillation can be eliminated if $T_s D_{i,i} < M_{i,i}$, which can only be achieved with a sufficiently small time step $T_s$ for typical $M_{i,i}$ and $D_{i,i}$. To allow larger $T_s$, we rewrite the Euler damping term $\bm{I}-T_s\bm{M}^{-1}\bm{D}$ using the closed-form integral as $\exp(-T_s\bm{M}^{-1}\bm{D})$. 

\subsection{Human Modelling}
\subsubsection{Human Forces}
The human is modelled stochastically as an impedance, where applied force depends on the pose and mode. For example, in an assembly task the modes represent different assembly goal locations where the part can be moved to (see Fig. \ref{fig:exp_setup}). The models are independent for each mode $n$ and are defined as
\begin{equation} \label{eq:human_model}
    {\bf \bm{f}^H} \sim p(\bm{f}^H|\bm{x}^*, n).
\end{equation}
This relationship is realized with Gaussian processes (GP) \cite{rasmussen2006}, where the human force is regressed over the position/orientation of the robot, where each mode has its own GP model. Each element of the human force is regressed in parallel, where the force data is treated independently.

The pose $\bm{x}^*$ reflects a change in orientation representation from $\bm{x}$. Rotation vectors are used to represent orientation for regression, where for a rotation vector $\bm{v}\in\mathbb{R}^3$, the direction describes the axis of rotation, and the magnitude the angle of rotation $\Vert \bm{v}\Vert = \theta$. This was found to have benefits over Euler angles and quaternions, which had discontinuities and were over-parameterized, respectively. The $*$ is dropped in the sequel, but all regression is done over rotation vectors, and admittance calculations over the TCP XYZ axis angles.

\subsubsection{Human Kinematics}
As some ergonomic costs are more naturally expressed in terms of human joint torques, we also consider a 4-DOF kinematic model of the human arm with three rotational DOF at the shoulder and one rotational joint for the elbow extension. This model has parameters of $l_1$ and $l_2$, for the length of the upper and lower arm, and $\bm{x}^{sh}\in\mathbb{R}^3$ for the human shoulder position. Using these parameters and the human joint angles $\bm{q}\in\mathbb{R}^4$, we can calculate the human hand position $\bm{x}^H\in\mathbb{R}^3$ with forward kinematics as
\begin{equation}
    \bm{x}^H =  \mathrm{FK}(\bm{q},\bm{x}^{sh}), \label{human_FK}\\
\end{equation}
Forces measured at the end-effector can be be translated into joint torques $\bm{\tau}$ as
\begin{equation}
    \bm{\tau} = \bm{J}_H(\bm{q})^T\bm{f}^H_l, \label{human_torque}
\end{equation}
where $\bm{f}^H_l$ are only the linear forces, and $\bm{J}_H(q)\in\mathbb{R}^{3\times 4}$ is the standard Jacobian matrix.


\subsection{Stochastic dynamics}
\begin{figure}
    \centering
    \includegraphics[width=0.7\columnwidth]{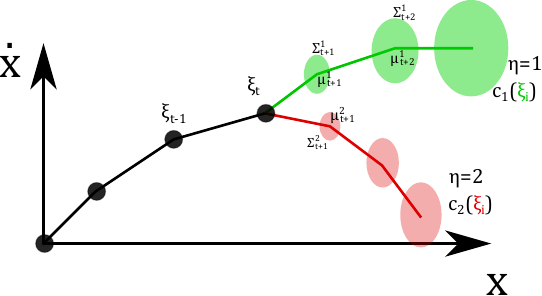}
    \caption{Multi-modal future trajectory, where each possible human mode induces a different course of the trajectory, and each possible trajectory also has uncertainty.}
    \label{fig:multimodal_traj}
    \vspace{-4mm}
\end{figure}

As the human model is stochastic and affects the robot trajectory, the future state trajectory will also be stochastic. Each mode induces a distribution on the future state, and when the mode is uncertain, the future trajectory will be a multi-modal distribution such as seen in Fig. \ref{fig:multimodal_traj}. 

Denoting the dynamics of \eqref{eq:FirstOrderDiscSysDyn} with state-space representation as $\bm{\xi}_{t+1} = \bm{A}_t\bm{\xi}_t + \bm{B}_t(\bm{f}^H_t - \bm{f}^R_t)$, the future trajectory in mode $n$ is distributed as $\bm{\xi}_t \sim \mathcal{N}(\bm{\mu}_t^n, \bm{\Sigma}_t^n)$, and the mode dynamics are defined by
\begin{eqnarray}
    \bm{\mu}^{H,n}_t, \bm{\Sigma}^{H,n}_t & = &\mathrm{GP}_n(\bm{\mu}^n_t, \bm{\Sigma}^n_t), \label{human_force} \\
    \bm{\mu}^n_{t+1} &=& \bm{A}_t\bm{\mu}^n_t+\bm{B}_t(\bm{\mu}^{H,n}_t-\bm{f}^R_t), \label{unc_dynamics_mean} \\
    \bm{\Sigma}^n_{t+1} &=& \bm{A}_t\Sigma^n_t\bm{A}_t^T + \bm{B}_t\bm{\Sigma}^{H,n}_t\bm{B}_t^T, \label{unc_dynamics_cov}
\end{eqnarray}
where $\bm{\mu}^{H,n}$ and $\bm{\Sigma}^{H,n}$ are the mean and covariance of the $n^{\mathrm{th}}$ GP model. 

When using \eqref{unc_dynamics_mean} and \eqref{unc_dynamics_cov} for prediction starting at time $t$, they are initialized with $\bm{\Sigma}^n_t=\bm{0}$, and $\bm{\mu}^n_t = \bm{\xi}_t$, as it is assumed that the current state of the robot is measured. 

\section{Inference and MPC}
The modelling method presented in Sec. \ref{sec:models} is then applied to the inference and planning problems. The proposed solutions to these two problems are described in the two subsections below.

\subsection{Inference}
Given a series of observations $\bm{\xi}_{1:t} = [\bm{\xi}_1,\dots,\bm{\xi}_t]$ and $\bm{f}^H_{1:t}$, the human mode can be estimated using the assumed human dynamics from \eqref{eq:human_model}. We examine a case where the mode has a fixed prior distribution, and a case where the distribution of mode transitions are known.

\subsubsection{Mode without transition}
If the mode is fixed during the task (e.g. one of several goals is active), but has an initial distribution of $p(n)$, the posterior is defined as
\begin{eqnarray}
\begin{split}
  p(n | \bm{\xi}_{1:t}, \bm{f}^H_{1:t}) & = & \frac{p(n)\prod_{i=1}^t p(\bm{f}^H_i | \bm{\xi}_i, n)}{\sum_{m=1}^N p(m)\prod_{i=1}^t p(\bm{f}^H_i | \bm{\xi}_i, m)} \label{inference_update}\\
   & \propto & p(\bm{f}^H_t | \bm{\xi}_t, n)p(n | \bm{\xi}_{1:t-1}, \bm{f}^H_{1:t-1}),
\end{split}
\end{eqnarray}
which can be recursively calculated. We apply a floor to the belief before normalization, i.e. $p(n | \xi_{1:t}, f^H_{1:t}) = \max \left(\underline{b}, p(n | \xi_{1:t}, f^H_{1:t})\right)$, with typical values of $\underline{b} = 1e-6$.

\subsubsection{Mode with transitions}
If the distribution of mode transitions are known, $p(n_{t+1} | n_t, \xi_t)$, the posterior can also be calculated as
\begin{eqnarray}
\begin{split}
   p(n_{t+1} | \bm{\xi}_{1:t}, \bm{f}^H_{1:t}) & \\ \propto  \sum_{n_{1:t}}& p(n_1)\prod_{i=1}^t p(\bm{f}^H_i | \bm{\xi}_i, n_t)p(n_{t+1}|n_{t}, \bm{\xi}_t) \\
   \propto \sum_{n_t} & p(\bm{f}^H_{t} | \bm{\xi}_t, n_t) p(n_{t+1} | n_t) p(n_t | \bm{\xi}_{1:t-1}, \bm{f}^H_{1:t-1})\nonumber.  \label{transitions}
\end{split}
\end{eqnarray}
This also allows an efficient recursive calculation of belief in a specific human mode.

\subsubsection{Transforming force to direction}
The GP model for force independently models each element in $\bm{f}$, $p(\bm{f}^h|\bm{\xi},n) = \prod_{i=1}^6p(f^h_i|\bm{\xi},n)$, where $f^h_i$ is the $i^{th}$ element of the force vector. The direct approach described in \eqref{inference_update} can be noisy, e.g., if one element has a very low probability this can render the total likelihood low. To account for that, we developed an ad-hoc similarity measure, where the pseudo-likelihood is calculated as
\begin{eqnarray}
    s(\bm{f}^H, \bm{\mu}^H, \bm{\Sigma}^H) &=& \beta\Vert \bm{\mu}^H\Vert_2\log(\frac{1}{2}(\bm{f}^H)^T\bm{\mu}^H+\frac{1}{2}) \nonumber\\ & & -\sum_{i=1}^6\log(\bm{\Sigma}[i,i]),
\end{eqnarray}
where $\beta$ weights the relative priority given to the direction matching (typical value, $0.05$). This ad-hoc rule updates more strongly when forces are larger, and the value is higher when the model and measured force point in a similar direction, and smaller when the uncertainty in the GP model is larger. The developed ad-hoc similarity measure was found to result in a smoother mode inference for co-manipulation, when it is used to replace $p(\bm{f}^H|\bm{\xi},n)$ in the belief update \eqref{inference_update}.

\subsection{MPC Problem}
MPC iteratively solves an optimal control problem, choosing decision variables that (locally) minimize a cost function, while imposing consistency of the model dynamics   \eqref{eq:FirstOrderDiscSysDyn}. The MPC implemented algorithm uses a multiple-shooting transcription with a problem statement of
\begin{equation}
\begin{array}{r@{}l}
    \bm{u}_{t:t+H} &{= \arg\min_{\bm{u}} J(b_{t},\bm{\xi}_{t})} \\
    \mathtt{s.t.} \,\, &\forall n \in [1,\dots,N], \tau \in [t,\dots,t+H-1]: \\
    &{\bm{\mu}^n_t = \bm{\xi}_t,\,\,\bm{\Sigma}^n_t = 0} \\
    &{\vert\bm{\mu}^n_{\tau+1}-f(\bm{\mu}^n_{\tau},\bm{u}_\tau)\vert \leq \varrho} \\ 
    &{\vert\bm{\Sigma}_{\tau+1}-g(\bm{\mu}^n_{\tau},\bm{\Sigma}^n_{\tau},\bm{u}_\tau)\vert \leq \varrho}\\
    &{\bm{u} \in U}
    \label{mpc_statement}
\end{array}
\end{equation}
where $H$ is the planning horizon, $U$ is the range of allowed inputs, $\varrho$ the slack for the continuity constraints (the inequality is applied element-wise), $f(\mu_{t},u_t)$ derives $\mu_{t+1}$ following \eqref{unc_dynamics_mean}, and $g$ following \eqref{unc_dynamics_cov}.

The constraints in \eqref{mpc_statement} are nonlinear, so an interior-point nonlinear optimization solver is used (details in \S\ref{sec:implementation}). While nonlinear, the GP models can be written to be automatically differentiated, allowing calculation of the gradient and Hessian of the objective and constraints, significantly improving the speed and stability of the optimization.

The MPC framework here allows for different choices of $\bm{u}$ and $\bm{J}$. Recall $\bm{x}^{sh}$ is the human shoulder position, $\bm{q}$ are the human joint angles, $\bm{\tau}$ are the human joint torques, $\bm{f}^H$ is the human force, $n$ is the mode, $\dot{\bm{x}}$ is the robot velocity, $\bm{f}^R$ is the robot desired force.  

\subsubsection{Decision variables $u$}
The decision variable $\bm{u}$ can include any of the following:
\begin{eqnarray}
\bm{f}^R_{t:t+H} & & \mathrm{Robot\,\,trajectory} \nonumber\\
\bm{\Delta}^M_t, \bm{\Delta}^B_t & & \mathrm{Change\,\,in\,\,robot\,\,impedance} \nonumber\\ 
\bm{x}^{sh}, \bm{q}_{t:t+H} & & \mathrm{Human\,\,shoulder\,\,and\,\,joint\,\,traj.}\nonumber 
\end{eqnarray}
If $\bm{x}^{sh}$ and $\bm{q}_{t:t+H}$ are included as decision variables, an additional constraint of $\bm{x}^H = T(\bm{\mu}_t^n)$ is added to the MPC problem, where $\bm{x}^H$ is from \eqref{human_FK} and $T$ represents the hand grasp location relative to robot TCP. 

\subsubsection{Stage cost function $c_n$}
The general stage cost function is defined as:
\begin{equation}
\begin{split}
    c_n(\bm{\mu}, \bm{\Sigma}, \bm{\mu}_H, \bm{\Sigma}_H, \bm{\tau}, \bm{u}) & = \\
      & \bm{\mu}^T\bm{Q}_\mu\bm{\mu} + \mathrm{tr}(\bm{Q}_\Sigma{\bm{\Sigma}}) + \\ 
      &  \bm{\mu}_H^T\bm{Q}_H\bm{\mu}_H + \mathrm{tr}(\bm{Q}_{\Sigma,H}\bm{\Sigma}_H)  + \\
          &  \bm{\tau}^T\bm{Q}_J\bm{\tau} + {\bm{u}}^T\bm{Q}_u\bm{u},
\end{split} \label{cost_fn}
\end{equation}
where $\bm{\mu}$ and $\bm{\Sigma}$ are mean and covariance for the predicted state $\bm{\xi}$ in mode $n$, $\bm{\mu}_H$ and $\bm{\Sigma}_H$ are the mean and covariance of predicted human forces from \eqref{human_force}, $\bm{\tau}$ are the human joint torques from \eqref{human_torque} and $\bm{Q}_{\cdot}$ are the related weight-matrices for each cost factor.

In some applications, replacing $\bm{\mu}_H^T\bm{Q}_H\bm{\mu}_H$ with $(\bm{\mu}_H+\bm{f}^R_t)^T\bm{Q}_H(\bm{\mu}_H+\bm{f}^R_t)$ can be more robust -- the mean human force goes to zero outside the training data (in zero-mean GPs, as used here), and a large penalty on only $\bm{\mu}^H$ can lead to trajectories which seek to leave the training data.   

\subsubsection{Total objective function $J(b,\xi)$}
The total objective sums the stage costs and considers the current belief $b_t$. Two varieties were used: the simple expectation over belief $J_E$, and risk-sensitive cost $J_R$, which is adapted from \cite{medina2012a}, where $\alpha$ adjusts the sensitivity to risk. The two cost functions are defined as
\begin{eqnarray}
     J_E = \sum_{t}^{t+H}\sum_{n=1}^N b_t[n] c_n(\bm{\xi}^n_t, \bm{f}^R_t, \bm{f}^{H,n}_t), \label{exp_cost}\\
     J_R = -2\alpha^{-1}\ln\mathbb{E}_{n\sim b_t} e^{\left(\sum_{t}^{t+H}-\frac{1}{2}\alpha c_n(\bm{\xi}^n_t,\bm{f}_t^R, \bm{f}_t^{H,n})\right)}. \label{risk_sensitive_cost}
\end{eqnarray}

\section{Implementation \label{sec:implementation}}
This section presents the overall integration and details on the implementation of the proposed method. The code, parameter settings, training data, and experimental results are available at \url{https://owncloud.fraunhofer.de/index.php/s/kmCZvlKOghclHy9}.

\subsection{Data collection and commissioning}
An overview of the commissioning and online execution is shown in Algorithm \ref{algo}.  To initialize the GPs, data must be collected for each mode. For co-manipulation tasks, this is done here by having the robot in a passive admittance control mode, where it acts as a mass-damper system. The default damping gains used are $\bm{M}=\mathrm{diag}([12, 12, 12, 1, 1, 1])$ where $\mathrm{diag}$ is the diagonalization operation, and $\bm{D}=\mathrm{diag}([1100, 1100, 1100, 200, 200, 200])$. All units are in SI standard units, and radians are used.
 
\SetKwComment{Comment}{}{}
\SetKwBlock{kwInit}{Initialize GPs}{}
\SetKwBlock{kwBel}{Belief Thread {\rm (50 Hz)}}{}
\SetKwBlock{kwCtl}{Control Thread{\rm ($\sim$5-20 Hz)}}{}
\SetKwBlock{kwOnline}{Online Execution}{}
\begin{algorithm}
\DontPrintSemicolon
\caption{Data collection and execution, with GP hyperparameters $\phi$ \label{algo}}
\kwInit{\par
  \For{n = 1 \dots N}{
    Collect data $\mathcal{D}_n \leftarrow \{\{\xi_1,f_1\},\dots\}$ with mass/damper admittance\;
    (optional) $\phi = \mathtt{fit\_hyperparams}(\mathcal{D}_n, \phi)$\;
    (optional) $\mathcal{D}_n = \mathtt{sparsify\_GP}(\mathcal{D}_n, \phi)$\;
    $\mathrm{GP}_n \leftarrow \mathtt{build\_GP}(\mathcal{D}_n, \phi)$\;
  }
}
\kwOnline{
\kwBel{\par
    $\xi, f \leftarrow \mathtt{read\_state}()$\;
    $b \leftarrow \mathtt{belief\_update}(b,\xi, f, \mathrm{GP})$ by \eqref{inference_update}\;
}
\kwCtl{\par
    $u \leftarrow \mathtt{MPC\_solve}(\xi, b, \mathrm{GP})$ by \eqref{mpc_statement}\;
}
}
\end{algorithm}

\begin{table*}[]
\caption{Comparison of MPC problem statements and computational cost \label{comp_cost}}
\centering
 \begin{tabular}{ c c c c c c | r r r} 
 \hline
 Full GP Cov. & State Cov. & \# GP Points & J & Imp Params & $x^{sh}$,$q_{t:t+H}$  & Cold (sec) & Avg Warm & Worst Warm \\ [0.5ex] 
 \hline\hline
 Yes & Yes & 50 & Expected & No  & No & 11.524 & 0.157 & 1.185 \\ 
 No & Yes & 50 &  Expected & No  & No & 5.023 & 0.062 & 0.215 \\ 
 No & No & 50 &  Expected & No & No  & 4.571 & 0.053 & 0.176 \\
 No & Yes & 35 &  Expected & No & No  & 5.239 & 0.066 & 0.225 \\  
 No & Yes & 50 & Expected & Yes & No  & 17.910 & 0.342 & 2.354 \\
 No & Yes & 50 & Expected & No & Yes  & 6.483 & 0.137 & 1.977 \\
 No & Yes & 50 & Risk Sens & No & No  & 44.223 & 0.330 & 2.399 \\
 [1ex] 
 \hline
 \end{tabular}
\end{table*}

\subsection{Software Implementation}
To enable stable and smooth performance of the designed controller, an MPC loop speed $>5$ Hz is targeted. The MPC problem is solved using CasADi \cite{andersson2019casadi} to interface with IPOPT \cite{wachter2006implementation} using the HSL MA57 linear solver \cite{HSL}. Solver parameters used for IPOPT are detailed in the above-referenced cloud, and warm start is used for all subsequent solves. The horizon and shooting nodes where set to $H=6$ with a sampling time $T_s = 0.10$ seconds. The continuity constraint slack was set at $\varrho = 10^{-5}$ meter, which is still within the limits of the robots precision. The maximum force applied by the robot $\bm{f}^R$ was constrained to be less than $20$ N and $6$ Nm for linear and rotational force. The state space was limited with regard to the maximum velocity, but the position of the robot was unconstrained within the MPC calculations. 

\begin{figure}
    \centering
    \includegraphics[width=0.48\columnwidth]{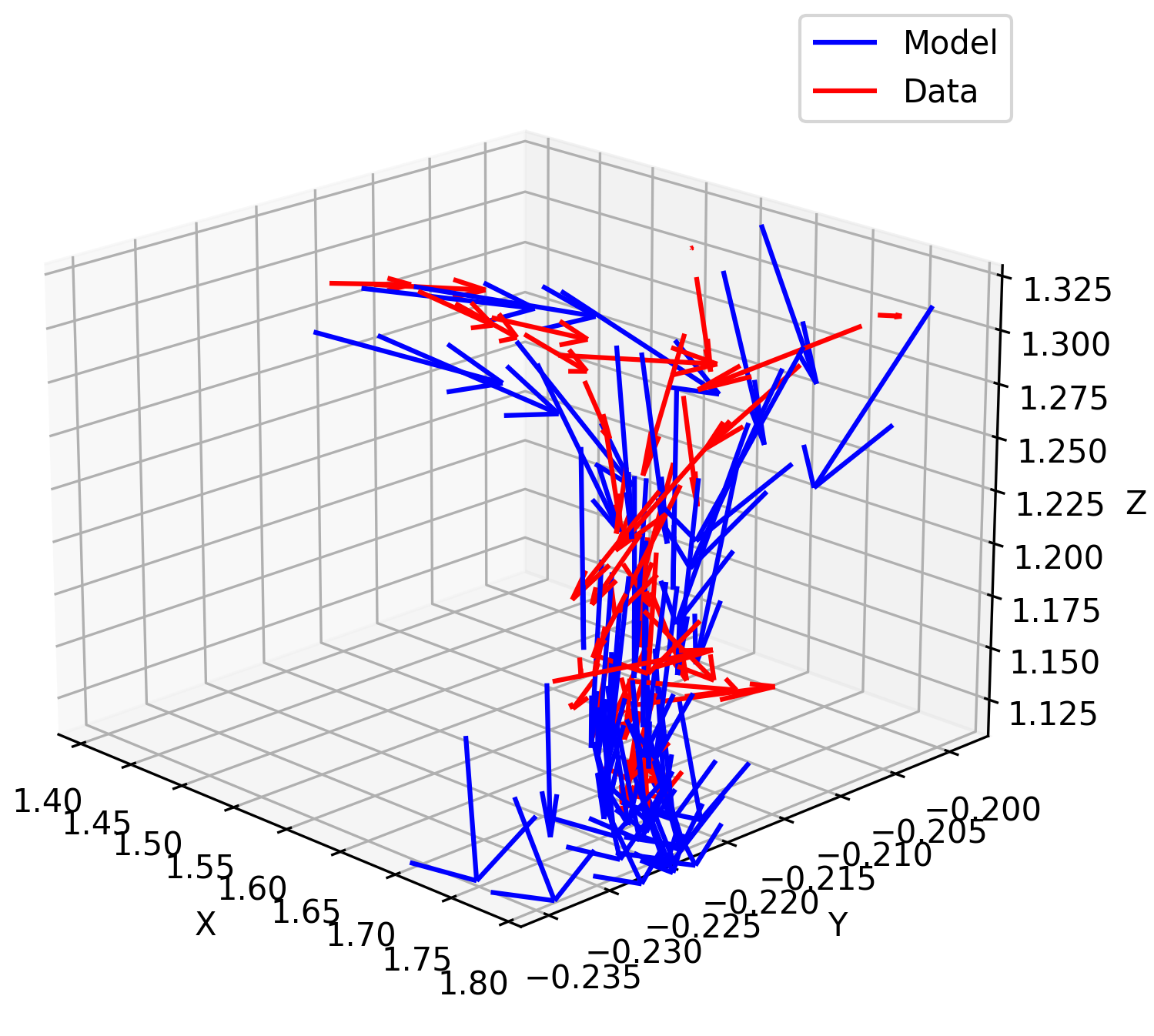}
    \includegraphics[width=0.48\columnwidth]{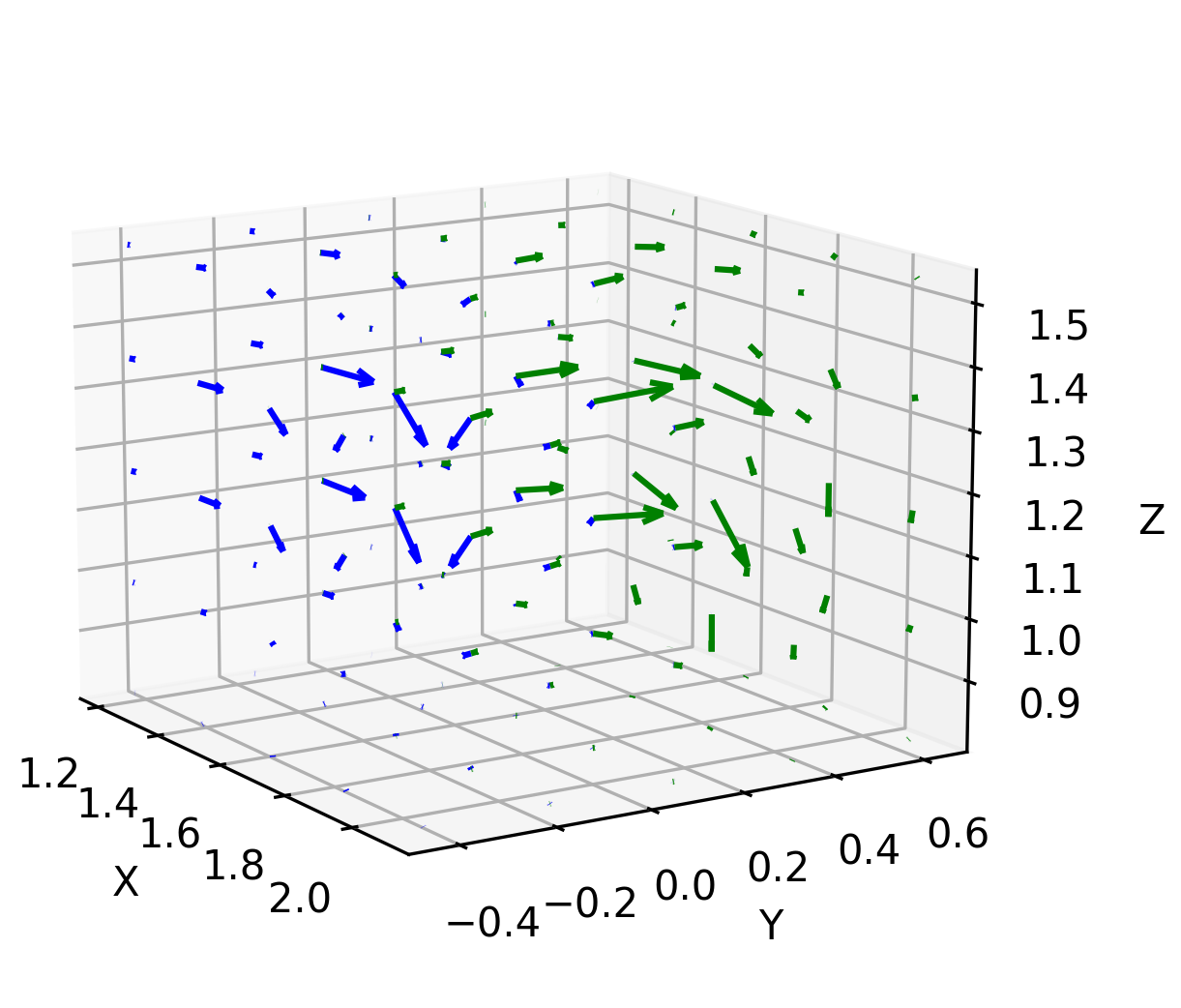}
    \caption{Linear force over Cartesian position: (Left) Training data and mean of the GP model evaluated near the training data. The contact forces from the pins can be seen as red arrows to the left/right, and are smoothed out by the GP model. (Right) mean force of two GP models with blue and green vectors.}
    \label{fig:gp_mixed}
    \vspace{-4mm}
\end{figure}

The GP regression we used builds on \cite{langaker2018} and is implemented in CasADi. This allows the automatic calculation of the gradient and Hessian for the interior point solver. A squared exponential GP kernel was used with 35-50 training samples, sub-sampled from three the demonstration trajectories, which started at different positions, as seen in the left of Fig. \ref{fig:gp_mixed}. An example of the forces fitted over position for two modes can be seen in Fig. \ref{fig:gp_mixed}, right.
Since GP model evaluations come with a high computational cost, the number of training samples necessary was reduced using Sparse GPs.

\subsection{Sparse GPs}
As every MPC loop involves at least $HN$ GP evaluations for each optimizer step and the asymptomatic complexity of a GP evaluation is $O(S^3)$, where $S$ is the number of training samples, the calculation time is highly sensitive to the number of training points. Since the collected data set contains thousands of samples, a method of sub sampling is needed. To address these two issues, sparse GPs are used, which can be calculated in $O(RS^2)$, where $R$ is the number of inducing variables that replace the original training data. Within sparse GPs the inducing variables are generated by minimizing the Kullback-Leibler divergence between a GP trained on a larger subset of the original data and an GP using artificial training data as described in  \cite{titsias2009variational}. In our experiments this method effectively reduced the amount of training samples -- $35$ samples had equivalent log likelihood to $50$ samples when subsampled by time.

\section{Experimental Validation}
\subsection{Hardware Validation}
The approach is validated on a large industrial robot, where a 16 Kg steel plate had to be manipulated and/or have objects mounted to it, as seen in Fig. \ref{fig:exp_setup}. This plate must be mounted on a set of two pins (insets of Fig. \ref{fig:exp_setup}) which have a loose running clearance fit (hole is oversized by $0.5$ mm), and there are two possible mounting positions for the plate. Here, we used $\bm{Q}_\mu=\mathrm{diag}([0,\dots,0,0.1,\dots,0.1)$ to penalize only velocity, $\bm{Q}_\Sigma=0.1\bm{I}$, $\bm{Q}_H = 0.1\bm{I}$, $\bm{Q}_{\Sigma,H} = 270\bm{I}$ --  $\bm{Q}_u=0.25\bm{I}$, and $\bm{Q}_{\Sigma,H}$ was found to be the most critical.

\begin{figure}
    \centering
    \includegraphics[width = 0.48\columnwidth]{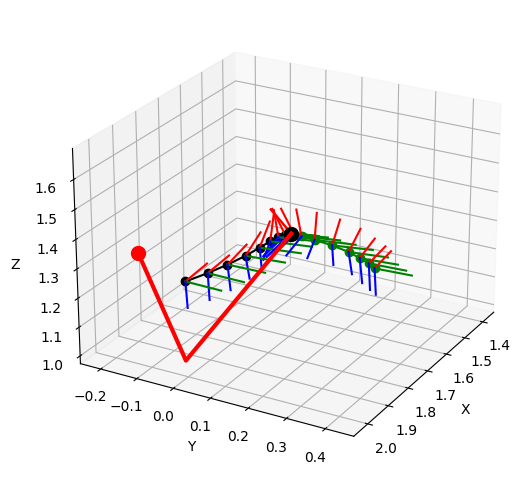} 
    \includegraphics[width = 0.48\columnwidth]{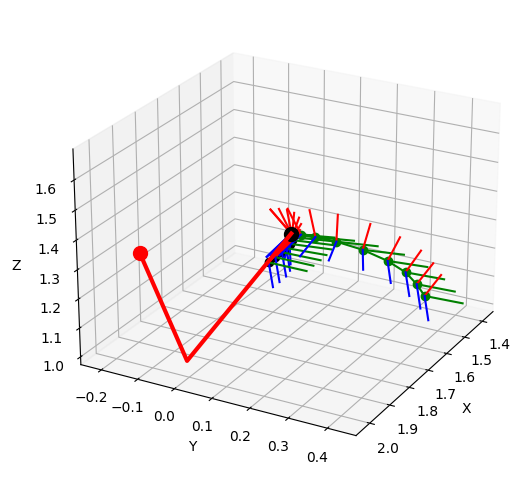}
    \caption{View of MPC planned trajectory for each mode (blue and green), with human arm model in red and current TCP location/orientation as black point. On the left, when the belief is $b_t = [0.5, 0.5]$ for the left and right goal, and on the right is $b_t = [0.05, 0.95]$. }
    \label{fig:mpc_plan}
    \vspace{-4mm}
\end{figure}

\begin{figure}
    \centering
    \includegraphics[width = \columnwidth]{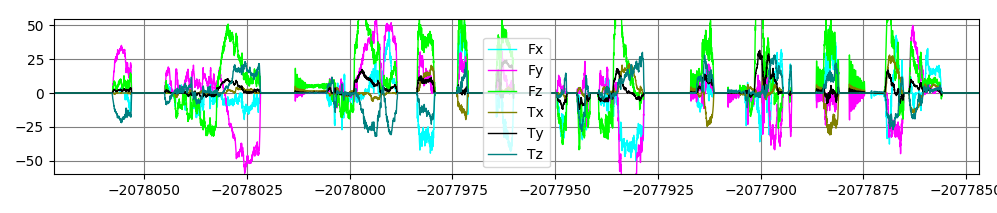} \\
    \includegraphics[width = \columnwidth]{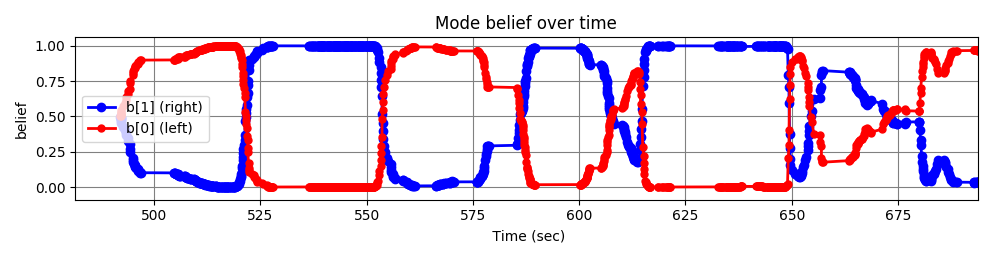} \\
    \includegraphics[width = \columnwidth]{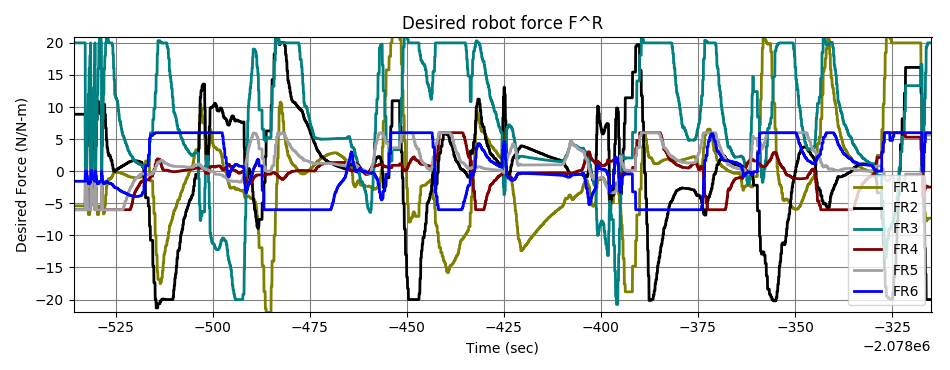} \\
    \caption{Time traces of applied forces (top), belief (middle) and desired force trajectory $F^R$ (bottom). }
    \label{fig:mpc_trajectories}
    \vspace{-4mm}
\end{figure}

Two MPC planned trajectories can be seen in Fig. \ref{fig:mpc_plan}, where only the belief is changed. With an even belief $b_t = [0.5, 0.5]$ (left), the planned forces $\bm{f}^R$ are smaller, whereas when the right belief is stronger (right), the MPC plans to move more in this direction.

The attached video shows the performance on physical hardware, and the resulting time plots can be seen in Fig. \ref{fig:mpc_trajectories}. The system is able to robustly detect changes in goal online, and provides appropriate active assistance. The system is also robust to a variety of starting positions, and can recover when the human perturbs the robot away from the goal, providing the flexibility of online motion planning. While the final accuracy is limited (+/- $2$ cm is typical), this fine positioning can be taken over by the human. Note that the goal position is not hard-coded, but is implicit from the demonstrations (human forces are smaller near the goal and model covariance is lower because more data is collected there). 

\subsection{Computational Cost}
To compare the relative computational cost of several aspects of the approach, the MPC loop is solved over recorded experiment data, allowing repeatable comparison while covering the state space. A computer with Intel i3-5010 @ 1.70 Ghz was used for all validation, with $H=5$ and $dt = 0.05$. If the `Full GP Cov.' option is false, the GP covariance is simplified, replacing the  $\mathrm{Tr}(\bm{Q}_{H,\Sigma}\bm{\Sigma_H})$ term in \eqref{cost_fn} with $\mathrm{Tr}(\bm{Q}_{H,\Sigma})\Sigma_{H,1,1}$, using just the first element (as the covariance kernel is shared between linear directions, and similar between linear and rotational). The `State Cov.' option defines whether or not the state covariance in \eqref{unc_dynamics_cov} is calculated. The cold start time (the first MPC solve, which includes building the MPC solver), average warm start time and worst-case warm start are compared.

It can be seen in Tab. \ref{comp_cost} that simplifying the full GP covariance makes a large computational difference -- this simplification does not reduce performance when similar hyperparameters are used between the DOF, as the covariance will be similar between DOF. Reducing the GP size and not calculating state covariance make minor improvements in average calculation time.  

Adding impedance parameters to the problem significantly increases both mean and worst-case calculation time. Similarly, adding the human joints/shoulder to the problem significantly increases the worst-case calculation. Using a risk-sensitive objective results in severe computational costs. These three problem statements are not currently feasible for online control in the current approach.

\subsection{Co-optimization of human joints and impedance trajectory}
The attached video shows the co-optimization of the human joint trajectory, where the co-optimized trajectory for human joints finds a solution that reduces the moment arm between point of force application and shoulder, thereby reducing the shoulder torques required. However, transcribing all of the ergonomic constraints and costs remains a topic for future work. Similar video results show the co-optimization of  impedance parameters, although the current cost functions and test applications typically result in simply minimizing the impedance parameters. 

\section{Conclusion}
The feasibility of MPC for direct physical HRI in 6 DOF has been shown, with both realistic data-collection requirements (3 demonstrations) and online computation speed (15 Hz). We note that while the approach considers a stochastic human model and dynamics, the largest functional advantage of this approach is in the human model, in keeping the robot near the training data. Considering uncertainty in the trajectory has not yet been shown to offer new functionality, although new use-cases and objectives may establish benefits. 

The proposed approach also has known limitations, many of which are shared with other approaches, but are listed for completeness. The approach does not distinguish between environmental forces and human forces, requiring care in contact tasks. GP covariance does not capture heteroscedastic properties (i.e. if there is larger variance in a region of the state space) -- modelling changes in underlying covariance requires splitting the GP, which is then computationally expensive. Fitting GP hyperparameters with log likelihood on limited data can be problematic, to address this we limited the range of hyperparameters or used fixed ones.

\ifdefined\ARXIV
\appendix
\section{Implementation Details}

\subsection{MPC Solver parameters}
IPOPT v3.14 was used, interfaced through CasADi v3.4.5. IPOPT has many (maybe too many) options that can be adjusted for the interior point solver algorithm. In the course of this work some IPOPT parameters were examined with regard to their influence on time and and jitter of computation. As a result some parameter were set on non-default values: The $\mu$-strategy used by the solver was set on 'adaptive' using probing instead of the default quality-function. Additionally the solver was set to expect an infeasible problem such that the restoration phase (RP) is activated sooner and the constraint violation is made smaller before leaving the RP. The constrain violation tolerance $\varrho$ was set within the limits of the robots precision to $10^{-5}$.  Using the HSL linear solvers, especially MA57 and MA86, significantly improved performance. 

\subsection{MPC Formulation}
The dynamics of \eqref{unc_dynamics_mean} and \eqref{unc_dynamics_cov} were implemented with for loops instead of matrices, as that better tracked the sparsity pattern of the problem (i.e. which variables are indepndent of each other, which helps speed up the interior point method). CasADi was used to autogenerate C code for the MPC problem, allowing a typical speedup of about $2-3$x, but often taking several minutes to compile. The $\mathtt{gcc}$ flags for $\mathtt{-Os}$ were used, other optimization flags often exceeded available memory.

\subsection{Gaussian Processes}
The force measurements were pre-processed by removing all data points with magnitude less than 3 N, then they were subsampled by time to reach the desired number of steps.  Between $35$ and $50$ data points were used per mode here.  

A squared exponential kernel is used, where $K(x,y) = \sigma_f^2\exp(\Vert x-y\Vert_2^2/l^2)$, with variance of observation noise $\sigma_n^2$.  While the hyperparameters can be optimized to maximize log likelihood, this often resulted in some parameters being extreme, especially in smaller datasets. Hard-coded hyperparameters were used, for linear DOF: $l=0.12$, $\sigma_f=2.75$, $\sigma_n=4.0$ and rotational DOF: $l=1.1$, $\sigma_f=0.95$, $\sigma_n=1.0$.

To keep computational costs of the GP down, we found some tricks, which are to a degree specific to the CasADi framework:
\begin{enumerate}
    \item Keep as few symbolic parameters in the automatic differentiation computation graph -- if the hyperparameters are first introduced as symbolic so they can be optimized, the GP should be re-built after optimization with the parameters hard coded. 
    \item In CasADi, ther is an option to expand symbolic expressions (doing inline substitutions).  This increases the memory footprint, but improved execution speed. 
    \item Using built-in parallelization for the execution of the covariance kernel function can help, in CasADi this is via the $\mathtt{map}$ command. 
\end{enumerate} 
\fi

\bibliographystyle{IEEEtran}
\bibliography{references}

\end{document}